\documentclass[runningheads]{llncs}
\usepackage[T1]{fontenc}
\usepackage{graphicx}
\usepackage{microtype}
\usepackage{booktabs}
\usepackage{multirow}
\usepackage{amsmath}
\usepackage{amssymb}

\usepackage{subcaption}
\usepackage{wrapfig}

\begin{document}
\title{Multimodal Event Detection: Current Approaches and Defining the New Playground through LLMs and VLMs}
%
%
\author{Abhishek Dey\inst{1} \and Aabha Bothera\inst{1}\and Samhita Sarikonda\inst{1} \and Rishav Aryan\inst{1} \and Sanjay Kumar Podishetty\inst{1}\and Akshay Havalgi\inst{2}\thanks{This work is not related to Akshay Havalgi's position at Amazon, and was completely outside of the regular duties of his position.} Gaurav Singh\inst{1}$^{\bigstar}$ \and Saurabh Srivastava\inst{1}$^{\bigstar}$ }
\authorrunning{Dey et al.}
%
\institute{George Mason University\and Amazon $^{\bigstar}$Equal Senior Authorship}
\maketitle              
\begin{abstract}
In this paper, we study the challenges of detecting events on social media, where traditional unimodal systems struggle due to the rapid and multimodal nature of data dissemination. We employ a range of models, including unimodal ModernBERT and ConvNeXt-V2, multimodal fusion techniques, and advanced generative models like GPT-4o, and LLaVA. Additionally, we also study the effect of providing multimodal generative models (such as GPT-4o) with a single modality to assess their efficacy. Our results indicate that while multimodal approaches notably outperform unimodal counterparts, generative approaches despite having a large number of parameters, lag behind supervised methods in precision. Furthermore, we also found that they lag behind instruction-tuned models because of their inability to generate event classes correctly. During our error analysis, we discovered that common social media issues such as leet speak, text elongation, etc. are effectively handled by generative approaches but are hard to tackle using supervised approaches.


\end{abstract}
\section{Introduction}
Natural disasters—such as earthquakes, hurricanes, wildfires, and floods—pose severe threats to human lives, infrastructure, and economies. The ability to detect and track such events in real-time is crucial for disaster response teams to issue timely warnings, coordinate relief efforts, and mitigate damage~\cite{ref_article1}. While traditional sources of disaster-related information, such as governmental reports and news articles, provide structured and validated updates, they often suffer from delays in reporting. In contrast, social media platforms offer real-time, high-volume, and publicly accessible streams of disaster-related data, making them invaluable for early event detection \cite{ref2}. However, extracting structured and actionable insights from social media is inherently challenging due to the informal nature of the text, high noise levels, and the multimodal nature of shared content (e.g., text, images, and videos).

Event Detection (ED)~\cite{ref3} in natural language processing (NLP) serves as the first step in structuring real-world event information by identifying instances of events in text. Classical ED research has focused on structured datasets such as ACE 2005 \cite{ref3}, which are predominantly composed of formal news articles. In these settings, ED is framed as the task of recognizing what happened, along with contextual information such as who was involved, when, where, and how. However, applying ED techniques to social media for disaster monitoring introduces significant challenges, including data sparsity, informal language variations, and the presence of multiple modalities.

Existing approaches that analyze social media disaster-related content primarily treat the problem as a classification task—determining whether a given post pertains to a disaster event or not—without explicitly detecting events in a structured manner. For instance, \cite{ref_article1} explored classification strategies to assess the challenges of multimodal disaster detection but did not frame the task within the ED paradigm. Such classification-based approaches lack interpretability and fail to capture finer-grained event attributes, such as location, time, and evolving event dynamics, which are critical for disaster response efforts.

In addition, social media presents unique linguistic and structural challenges that further complicate event detection. Unlike well-formed text in news articles, social media posts frequently exhibit informal writing styles, including leet speak (e.g., ``gr8 fl00d''), elongated words (e.g., ``shaaaking!!!''), inconsistent capitalization, abbreviations, and code-mixing. These variations introduce noise, making it difficult for traditional ED models trained on well-structured text to generalize effectively. In our study, we systematically analyze the impact of these linguistic perturbations on event detection and evaluate the robustness of modern transformer-based models, including BERT-based event extraction models. Our findings indicate that while current ED models struggle with these variations, large language models (LLMs) such as GPT-based models, exhibit significantly better adaptability, particularly in identifying events from highly distorted social media text.

To the best of our knowledge, this is the first study to investigate multi-modal ED on social media for natural disasters while simultaneously analyzing the linguistic challenges inherent to such datasets. Unlike prior classification-based approaches, we frame the task as structured ED, incorporating both textual and visual signals to improve robustness. Moreover, we provide an in-depth analysis of how informal language and social media writing styles impact ED performance, identifying four to five common perturbations that remain unresolved by existing ED models but can be effectively addressed by LLMs.

To summarize, our key contributions are as follows:
\begin{enumerate}
    \item  We introduce a novel formulation of multi-modal ED for natural disasters, moving beyond prior classification-based methods to structured event detection.
    \item We conduct a comprehensive study of linguistic perturbations in social media disaster-related posts, identifying major challenges that hinder existing ED models.
    \item We evaluate traditional transformer-based ED models (e.g., BERT-based models) on social media texts and highlight their limitations in handling informal language variations.
    \item We demonstrate that GPT-based LLMs perform poorly compared to traditional ED models in recognizing events under social media-specific noise. Our findings resonate with the recent studies on ED which also highlights the limitation of large models compared to smaller supervised approaches, offering insights into their adaptability for real-world event monitoring.
    \item We establish a benchmark for multi-modal ED in social media disaster scenarios, providing a new evaluation framework to facilitate future research in this direction \footnote{Our code is available at \url{https://github.com/salokr/MultiModelEventDetection}}.
\end{enumerate}

\section{Problem Formulation}
\paragraph{\textbf{Supervised Approaches.}}

Our objective is to develop a multi-modal event detection system that effectively integrates textual and visual data from social media to identify and categorize disaster-related events. This system is designed for real-time deployment to facilitate timely and accurate information dissemination during disaster scenarios.

\begin{figure}[t!]
    \centering
    \includegraphics[width=.5\linewidth]{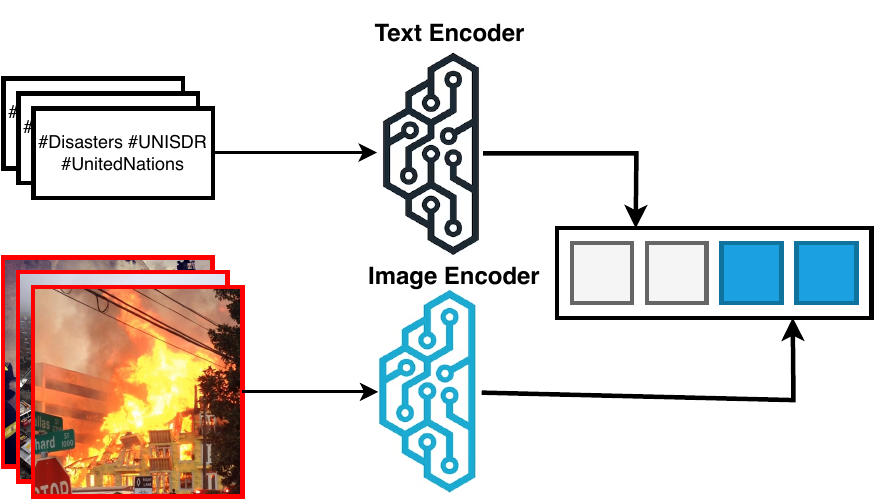}
    \caption{Overview of Multi-modal Event Detection. In addition to text, the input is accompanied by images and are fed to model to identify events.}
    \label{fig:overview}
\end{figure}

The input to our system (Fig. \ref{fig:overview}) consists of two primary modalities:
\begin{enumerate}

    \item \textbf{Textual Description ($T$)}: A sequence of token embeddings extracted from a pre-trained language model, capturing the linguistic and contextual nuances of social media posts.

    \item \textbf{Image ($I$)}: A set of visual features obtained from a deep vision model, providing complementary context to textual information for more robust event identification.

\end{enumerate}

Given an image $I$ and its corresponding textual description $T$, the goal is to classify the multi-modal input into a predefined set of disaster event categories. Formally, we define this task as learning function $f$ that maps the textual and visual modalities to a set of disaster-related categories:
\[
f : (I, T) \rightarrow C
\]
where $C$ represents a predefined taxonomy of disaster event categories, such as \textit{``flood'', ``fire'', ``human damage'', ``damaged nature'', and ``damaged infrastructure''}. 

Our model is trained under a supervised learning framework, using annotated disaster-related social media data. The classification function $f$ is optimized to maximize the predictive performance across both modalities while addressing challenges inherent to social media text, such as informal language, abbreviations, and linguistic perturbations. A high-level overview of our approach is illustrated in Fig. \ref{fig:overview}.

\paragraph{\textbf{Prompting Approaches.}}
In recent years, Vision-Language Models (VLMs) have demonstrated remarkable capabilities in understanding and integrating visual and textual data, leading to significant advancements in various tasks, including image classification, object detection, and event detection. Their ability to learn from vast amounts of image-text pairs enables them to perform zero-shot and few-shot learning, making them particularly valuable in scenarios with limited labeled data. 
\begin{figure}[h!]
    \centering
    \includegraphics[width=\linewidth]{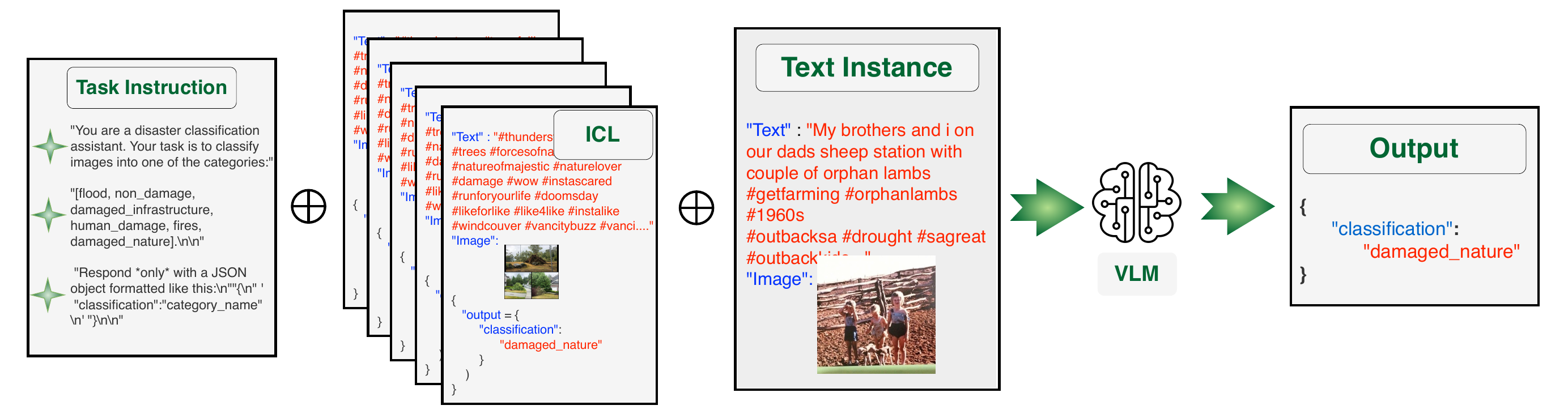}
    \caption{Overview of few-shot prompting strategy to perform ED. In addition to providing a few-shot in-context demonstration (ICL), we also provide the model with task instruction and output format to extract the correct event class for the test instance.}
    \label{fig:overview_pp}
\end{figure}
Prompting has emerged as an effective technique to adapt pre-trained models to specific tasks with minimal labeled examples. By crafting task-specific prompts, models can be guided to generate more accurate and contextually relevant outputs. In the context of VLMs, prompting facilitates the alignment of visual and textual modalities, enhancing the model's performance in few-shot settings.

To enable effective multi-modal event detection using few-shot prompting, we construct a five-shot image-and-text-based prompt to guide the model towards structured event identification. The prompt consists of five exemplar instances, each comprising:
\begin{enumerate}
    \item \textbf{Image ($I$)}: A disaster-related image.
    \item \textbf{Text ($T$)}: A social media post describing the disaster.
    \item \textbf{Event Category ($C$)}: The ground-truth label mapping the instance to a predefined disaster category.
\end{enumerate}
Our input sequence is prepared as:

[$\$$-task\_instruction $||<$Example-1$>||\dots||<$Example-5$>\$-$test\_instance] where ``$\$$-task\_instruction'' is the natural language task-instruction as shown in Fig. \ref{fig:overview_pp} and $\$-$test\_instance represents the test instance for which few-shot inference is performed. To facilitate automated evaluation, we instruct the models to output their predictions in a structured JSON format, enabling systematic answer extraction and quantitative analysis.

\section{Related Works}

\paragraph{\textbf{Event Detection (ED).}}
ED has been extensively studied in text-based data, particularly in the domain of news \cite{ref3,ref4,ref5,ref6}. While early approaches primarily relied on syntactic and semantic features, recent neural models have improved event extraction through contextualized embeddings and joint modeling techniques. Additionally, leveraging multimedia features has been shown to enhance text-based event extraction \cite{ref7,ref8,ref9,ref10}. In the domain of visual event extraction, events are often formulated as simpler actions or activities \cite{ref11,ref12}, typically within everyday life contexts. However, these works either focus on purely text-based ED or visual event classification, whereas our study explores \textbf{multi-modal event detection} by integrating textual and visual signals, particularly leveraging generative models such as VLMs and LLMs.

\paragraph{\textbf{Multi-Modal Event Detection (ED).}}
Recent multi-modal event detection methods primarily focus on aligning images and text for a unified event representation \cite{ref15}. However, these methods often neglect structural and semantic roles in event representation. UniVSE \cite{ref16} incorporates entity attributes and relations but lacks a fully structured event modeling approach. Recent advances using pre-trained vision-language models \cite{ref17} have improved multi-modal event extraction by integrating visual context, yet they largely formulate ED as a classification task constrained by predefined event schemas, limiting generalization. Furthermore, most existing methods assume that text-image pairs are always available, making them less effective in real-world disaster scenarios where one modality may be missing or incomplete \cite{ref18}. Additionally, situation recognition (SR) models predict events and arguments from images \cite{ref19} but do not localize arguments in a structured manner. In contrast, our approach redefines ED as a generative problem, allowing for flexible handling of missing modalities while providing structured event representations.

\paragraph{\textbf{Vision-Language and Large Language Models (VLMs/LLMs).}}
With the rapid advancement of multi-modal AI, systems can now process and integrate diverse modalities such as text, images, speech, and gestures, expanding the capabilities of AI applications. Vision-Language Models (VLMs) have been widely used in tasks such as Visual Question Answering (VQA), image captioning, and text-image retrieval, demonstrating significant improvements in multi-modal understanding \cite{ref20}. These models leverage large-scale vision-text pretraining to align visual and textual representations effectively. Similarly, Large Language Models (LLMs) have transformed natural language processing, exhibiting strong generalization across various tasks, including translation, summarization, and question answering. Advanced transformer-based architectures allow LLMs to process vast amounts of textual data, making them highly adaptable for a wide range of applications.

\paragraph{\textbf{Our Approach.}}  
While prior works have explored event detection in both text-based and multi-modal settings, our work differs in several key aspects:
\begin{itemize}
    \item \textbf{Generative ED with VLMs and LLMs:} Unlike existing approaches that frame ED as a classification problem, we formulate it as a generative task, allowing our model to flexibly infer missing event attributes and adapt to unseen disaster scenarios.
    \item \textbf{Handling Missing Modalities:} Previous works assume the availability of both text and image inputs, whereas our method explicitly models scenarios where one modality is absent (e.g., text-only or image-only posts).
    \item \textbf{Analysis of Social Media-Specific Linguistic Perturbations:} We systematically study the effects of informal language variations (e.g., leetspeak, word elongation, code-mixing) on ED performance—an aspect overlooked in prior multi-modal ED studies.
    \item \textbf{Few-Shot Prompting for Multi-Modal ED:} Our work is the first to explore prompt-based multi-modal ED using VLMs, demonstrating their effectiveness in disaster-related event detection.
\end{itemize}

By redefining multi-modal event detection as a generative task, incorporating robust prompting strategies, and analyzing social media-specific linguistic challenges, our work paves the way for ED research in disaster scenarios.

\section{Experimental Setup}
In this section, we discuss the approaches, dataset, and evaluation metrics for experiments.
\paragraph{\textbf{Dataset.}} We utilize the disaster event dataset proposed by \cite{ref_article1}. This dataset includes a collection of social media posts, specifically tweets, along with their corresponding images. The dataset is tailored for disaster event identification and contains a variety of event classes relevant to our study. The dataset distribution is detailed in Table \ref{tab:dataset_statistics}.

\begin{table}[h!]
\centering
\label{tab:dataset_statistics}
\resizebox{.5\linewidth}{!}{%
\begin{tabular}{@{}lcccc@{}}
\toprule
Label & Test & Train & Total \\ 
\midrule
Non-damage & 888 & 2069 & 2957 \\
Damaged Infrastructure & 417 & 973 & 1390 \\
Damaged Nature & 155 & 359 & 514 \\
Flood & 116 & 268 & 384 \\
Fires & 104 & 242 & 346 \\
Human Damage & 72 & 168 & 240 \\
\midrule
Total & 1752 & 4079 & 5831 \\
\bottomrule
\end{tabular}
}
\caption{Dataset Statistics. We use the samples for which both the tweets and images exist.}
\label{tab:dataset_statistics}
\end{table}

The dataset covers a range of disaster types, including \textit{``flood'', ``fire'', ``human damage'', ``damaged nature'', and ``damaged infrastructure''}.

\paragraph{\textbf{Evaluation Metrics.}} Following the previous literature on event detection, we use precision, recall, and F-score as evaluation metrics \cite{ref3,ref5,ref25}.

\section{Approaches}
\label{app:approaches}
\subsection{Text-Only Model}
Formally, the unimodal BERT \cite{ref21} classifier is a text-only model that solely relies on the textual description $T$ to classify events. It leverages the BERT architecture to capture the intricate linguistic patterns within the text and produces a contextual representation $\mathbf{W}$. This representation is passed through a linear layer to obtain the final classification $\hat{y}$

\begin{equation}
\hat{y} = \text{softmax}(\mathbf{W}_{t}\mathbf{W} + \mathbf{b}_{t})
\end{equation}
where $\mathbf{W}_{t} \in \mathbb{R}^{D_{W} \times |\mathcal{C}|}$ and $\mathbf{b}_{t} \in \mathbb{R}^{|\mathcal{C}|}$ are the parameters of the text classification layer.
 Notably, in addition to BERT, we also experiment with ModernBERT \cite{ref22} to study the effect of the modern version of Language Models on ED tasks.

\subsection{Vision-Only Model}
This unimodal approach utilizes only visual data and is realized through the Unimodal EfficientNet Classifier. The model harnesses the prowess of EfficientNet to extract rich and descriptive visual features $\mathbf{V}$ from the image $I$. These features are subsequently fed into a linear classification layer to predict the event category $\hat{y}$.
\begin{equation}
\hat{y} = \text{softmax}(\mathbf{W}_{v}\mathbf{V} + \mathbf{b}_{v})
\end{equation}
where $\mathbf{W}_{v} \in \mathbb{R}^{D_{V} \times |\mathcal{C}|}$ and $\mathbf{b}_{v} \in \mathbb{R}^{|\mathcal{C}|}$ denote the learnable parameters of the visual classification layer. 

We experiment with EfficientNet-B3 and EfficientNet-B7 as unimodal visual encoders to evaluate the impact of varying visual model complexities on event detection performance.

\subsection{Vanilla Multimodal Fusion}
The first vision-language fusion baseline that we considered involves combining representations from BERT and EfficientNet in a vanilla fusion approach. Formally,
the fusion-based multimodal classifier (MC) aims to integrate image and text representations for event detection. The image model utilizes EfficientNet, a CNN known for its efficiency and efficacy, to extract visual features $\mathbf{V} \in \mathbb{R}^{D_{V}}$ from the input image $I$. Concurrently, the text model employs BERT to derive contextual embeddings $\mathbf{W} \in \mathbb{R}^{D_{W}}$ from the textual input $T$. These features are transformed to a common dimensional space and merged to form a joint representation $\mathbf{M} \in \mathbb{R}^{D_{M}}$, which is processed by a fusion layer to predict the event label $\hat{y} \in \mathcal{C}$.
\begin{equation}
\mathbf{M} = [\mathbf{V}; \mathbf{W}], \quad \hat{y} = \text{softmax}(\mathbf{W}_{f}\mathbf{M} + \mathbf{b}_{f})
\end{equation}
where $[\cdot;\cdot]$ denotes the concatenation operation, and $\mathbf{W}_{f} \in \mathbb{R}^{D_{M} \times |\mathcal{C}|}$ and $\mathbf{b}_{f} \in \mathbb{R}^{|\mathcal{C}|}$ are the learnable parameters of the fusion layer.

\subsection{Vanilla Multimodal Fusion (ModernBERT and ConvNeXt-V2)}
To leverage the representational power of state-of-the-art architectures, we design a cross-attention-based fusion model that combines ModernBERT for text encoding and ConvNeXt V2 \cite{ref23} for image encoding. This model aims to perform fine-grained multimodal fusion for event classification by learning aligned representations across the vision and language modalities.
For the visual stream, we use ConvNeXt V2, a high-performing convolutional architecture, to extract deep image features $\mathbf{V} \in \mathbb{R}^{D_V}$ from an input image $I$. For the textual stream, ModernBERT \cite{ref22} is employed to generate contextualized embeddings $\mathbf{W} \in \mathbb{R}^{D_W}$ from a given text $T$.
Both feature vectors are projected into a shared representation space of dimension $D_M$ and integrated using a cross-attention mechanism that enables direct interaction between the image and text features. The resulting fused representation $\mathbf{M} \in \mathbb{R}^{D_M}$ is passed through a classification head to predict the event label $\hat{y} \in \mathcal{C}$.
\begin{equation} \mathbf{M} = \text{CrossAttention}(\mathbf{W}', \mathbf{V}'), \quad \hat{y} = \text{softmax}(\mathbf{W}{f}\mathbf{M} + \mathbf{b}{f}) \end{equation}
Here, $\mathbf{W}' = \text{ReLU}(\mathbf{W}{\text{text}}\mathbf{W})$ and $\mathbf{V}' = \text{ReLU}(\mathbf{W}{\text{img}}\mathbf{V})$ are the projected text and image features, respectively. The cross-attention module learns to align relevant textual and visual patterns. $\mathbf{W}{f} \in \mathbb{R}^{D_M \times |\mathcal{C}|}$ and $\mathbf{b}{f} \in \mathbb{R}^{|\mathcal{C}|}$ are trainable classification weights and biases.
The entire model is fine-tuned end-to-end with a cosine learning rate scheduler and early stopping strategy. During training, a class-balanced sampling strategy is employed to mitigate class imbalance using a weighted random sampler derived from inverse class frequency.

\subsection{Dual Attention Fusion}
The attention mechanism computes a set of weights $\alpha \in \mathbb{R}^{N \times (H \times W)}$, which are used to modulate the feature representations based on the interactions between the modalities. The attended features from the image and text are given by:

\begin{equation}
        \mathbf{I}_{att} = \text{Attention}(\mathbf{Q}_t, \mathbf{K}_i, \mathbf{V}_i)
\end{equation}
\begin{equation}
    \mathbf{T}_{att} = \text{Attention}(\mathbf{Q}_i, \mathbf{K}_t, \mathbf{V}_t)
\end{equation}

where $\mathbf{Q}_t$, $\mathbf{K}_i$, $\mathbf{V}_i$ denote the query, key, and value for the textual data, and $\mathbf{Q}_i$, $\mathbf{K}_t$, $\mathbf{V}_t$ for the image data, respectively. The attended features are then concatenated and passed through a fusion layer to predict the event category:
\begin{equation}
    \hat{c} = \text{softmax}(\mathbf{W}_f[\mathbf{I}_{att}; \mathbf{T}_{att}] + \mathbf{b}_f)
\end{equation}
where $\mathbf{W}_f$ and $\mathbf{b}_f$ are the weights and biases of the fusion layer, and $\hat{c}$ is the predicted probability distribution over the event categories.

\subsection{Cross-Modal Attention}
The cross-modal Attention Classifier (CMAC) introduces an attention mechanism that dynamically aligns and enhances the textual and visual features based on their inter-modal relevance. Given the feature representations $\mathbf{V}$ and $\mathbf{W}$ extracted from the image $I$ and text $T$ respectively, the model computes cross-modal attention weights $\alpha_{c}$ which are used to refine the features. The enhanced features $\mathbf{V'}$ and $\mathbf{W'}$ are concatenated to form the multimodal representation $\mathbf{M'}$, which is utilized for event classification.
\begin{align}
\alpha_{c} &= \text{softmax}(\mathbf{Q}_{W}\mathbf{K}_{V}^{\top}), \\
\mathbf{V'} &= \sum \alpha_{c} \odot \mathbf{V}, \quad \mathbf{W'} = \sum \alpha_{c} \odot \mathbf{W}, \\
\mathbf{M'} &= [\mathbf{V'}; \mathbf{W'}], \quad \hat{y} = \text{softmax}(\mathbf{W}_{c}\mathbf{M'} + \mathbf{b}_{c})
\end{align}
where $\mathbf{Q}_{W} \in \mathbb{R}^{D_{W} \times D_{A}}$, $\mathbf{K}_{V} \in \mathbb{R}^{D_{V} \times D_{A}}$, and $\odot$ denotes the element-wise multiplication. The parameters $\mathbf{W}_{c} \in \mathbb{R}^{D_{M'} \times |\mathcal{C}|}$ and $\mathbf{b}_{c} \in \mathbb{R}^{|\mathcal{C}|}$ are specific to the cross-attention fusion layer.

We experiment with GPT-4 as text-only LLM for event detection (\textbf{GPT-4(text only)}). For a fair comparison, we prompt LLaVA with only task instruction, class definition, and image to implement an image-only VLM baseline (\textbf{LLaVA (image-only)}). Finally, we experiment with the tweet and image with both GPT-4o (\textbf{GPT-4o (multi-modal))} and LLaVA (\textbf{LLaVA (multi-modal)}) for multi-modal comparison.

\subsection{Hyperparameters}
For generative approaches such as LLaVA and GPT-4, we set the temperature and top\_k values to .7 and .8, respectively. For linear layers, we use 1024 neurons and the large models of BERT and EfficinetNet.

\section{Results and Discussions}
We present our findings in Table 2, illustrating the performance metrics for each model. The results lead to several important observations about the efficacy of unimodal versus multimodal and generative approaches in disaster event detection.

\begin{table}[h!]
\centering
\resizebox{.9\textwidth}{!}{
\begin{tabular}{@{}lccc@{}}
\toprule
Model & Precision & Recall & F1-score \\ \midrule
\multicolumn{4}{c}{Supervised Approaches}\\\midrule\midrule

ModernBERT-large & 0.9067 & 0.9081 & 0.9070 \\
ConvNeXt V2 & 0.9229 & 0.9235 & 0.9227 \\
Multimodal Fusion 
(ModernBERT + ConvNeXt V2) & \textbf{0.9452} & \textbf{0.9467} & \textbf{0.9459
} \\ \midrule

BERT-only & 0.2483 & 0.4983 & 0.3314 \\
EfficientNet & 0.4700 & 0.4900 & 0.4798 \\
Dual 
Attention & 0.7640 & 0.7480 & 0.7559 \\
Cross-Modal Attention & 0.7815 & 0.7757 & 0.7686 \\
Multimodal Fusion & \textbf{0.9289} & \textbf{0.9247} & \textbf{0.9250} \\ \midrule
\multicolumn{4}{c}{Generative Approaches}\\\midrule\midrule
GPT-4o (text) & 0.3680 & 0.3558 & 0.3518 \\
GPT-4o (image) & 0.4620 & 0.4700 & 0.4660 \\
GPT-4o (multi-modal)& 0.6460 & 0.6300 & 0.6378 \\
LLaVA (image-only)& 0.4340 & 0.4470 & 0.4404 \\
LLaVA (multi-modal)& 0.5670 & 0.5740 & 0.5705 \\
\bottomrule
\end{tabular}
}
\label{tab:performance_metrics}
\caption{Performance Metrics for Different Models}
\end{table}
\vspace{-20pt}
\paragraph{\textbf{Effectiveness of Unimodal Approaches.}}
The performance of unimodal models highlights significant differences in how text-only and image-only models handle event detection. Among supervised methods, \textbf{ModernBERT-large (text-only)} achieves an F1-score of \textbf{0.9070}, while \textbf{ConvNeXtv2 (image-only)} slightly outperforms it with an F1-score of \textbf{0.9227}. This suggests that visual features provide stronger event cues than textual embeddings alone when dealing with disaster-related data.  For baseline models, \textbf{BERT-only} performs poorly, with an F1-score of \textbf{0.3314}, demonstrating that traditional transformer-based models struggle with event classification when trained solely on textual input. Similarly, \textbf{EfficientNet (image-only)} also underperforms, achieving an F1-score of \textbf{0.4798}, indicating that while images contain critical event information, they alone are insufficient for robust classification.  

Generative approaches perform significantly worse in unimodal settings. GPT-4o (text-only) achieves only 0.3518 F1-score, while GPT-4o (image-only) reaches 0.4660, demonstrating that VLMs, despite their large-scale pretraining, still struggle with precise event categorization when relying on a single modality. Similarly, LLaVA (image-only) performs suboptimally (0.4404 F1-score), showing that visual-only generative models are highly sensitive to ambiguous or missing contextual cues.  Overall, unimodal models exhibit limited effectiveness, with image-based models performing slightly better than text-based counterparts in supervised settings. However, generative approaches struggle significantly when constrained to a single modality, reinforcing the necessity of multi-modal integration for improved event detection.

\paragraph{\textbf{Advantages of Multimodal Approaches.}}
The results clearly demonstrate the superiority of multimodal models over their unimodal counterparts in both supervised and generative settings. The best-performing model, \textbf{ModernBERT + ConvNeXtv2}, achieves an F1-score of 0.9459, outperforming both its text-only (\textbf{0.9070}) and image-only (\textbf{0.9227}) components. This improvement underscores the fact that combining textual and visual information leads to more accurate and robust event detection.  Similarly, other multimodal supervised approaches such as \textbf{Multimodal Fusion} (BERT + EfficientNet) achieve an F1-score of 0.9250, significantly outperforming BERT-only (0.3314) and EfficientNet (0.4798). The Cross-Modal Attention model (0.7686 F1-score) also surpasses its unimodal counterparts, reinforcing that joint modeling of textual and visual data enables deeper contextual understanding.  

For generative approaches, the multi-modal versions of both GPT-4o and LLaVA significantly outperform their unimodal variants. \textbf{GPT-4o (multi-modal) achieves 0.6378 F1-score}, a considerable improvement over both GPT-4o (text-only, 0.3518) and GPT-4o (image-only, 0.4660). Similarly, \textbf{LLaVA (multi-modal) reaches 0.5705 F1-score}, showing a meaningful boost over LLaVA (image-only, 0.4404). This highlights that generative models struggle in unimodal settings but benefit significantly from cross-modal reasoning.  

In summary, multimodal approaches consistently outperform unimodal ones across all settings. The results indicate that disaster-related event detection requires both textual and visual information to resolve ambiguities and improve classification accuracy. In particular, generative models demonstrate a strong reliance on multi-modal integration, whereas supervised approaches further benefit from explicit fusion techniques, leading to state-of-the-art performance.

\paragraph{\textbf{Challenges with Generative Approaches.}}
As depicted in Table 2, generative approaches including GPT-4o and LLaVA fall behind their supervised counterparts. Qualitatively, we found that the information fabrication and their inability to follow task instructions are one of the major reasons for their performance. For example, instead of predicting the class \texttt{``Flood''}, LLaVA generated \texttt{``The input shows a flooding event''}. In addition, for non-disaster events, instead of generating ``\texttt{Non-Damage}'' class GPT-4 generates ``\texttt{There are no events in the input.}'' Notably, these instructions were not part of input instructions and hence can be categorized as information fabrication and mismatch.

Intriguingly, generative approaches were able to handle noisy text input such as text elongation and leet speaks. For example, GPT-4o was able to identify Leetspeaks like ``\texttt{4 people d!ed due to fire.}'' contains information related to a fire event that resulted in the death of 4 people. In addition, it can also handle cases such as typos and text elongation---a commonly seen writing style on social media. In summary, while these approaches were outperformed by supervised counterparts but also could pave the way for robust text processing resulting in strong multi-modal performance.

\begin{figure}[t!]
    \centering
    \begin{subfigure}{0.3\textwidth}
        \centering
        \includegraphics[width=\textwidth]{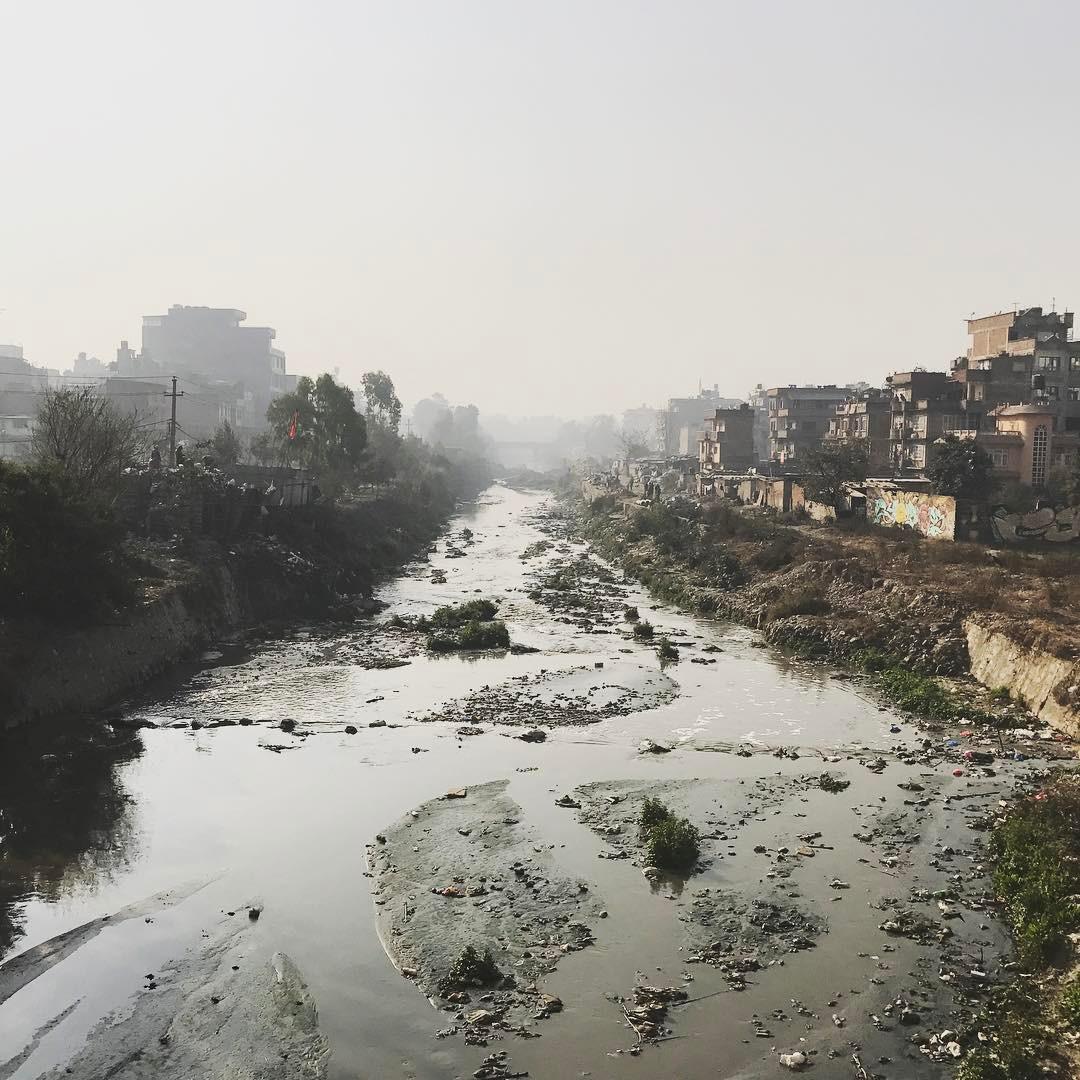}
    \end{subfigure}
    \begin{subfigure}{0.24\textwidth}
        \centering
        \includegraphics[width=\textwidth]{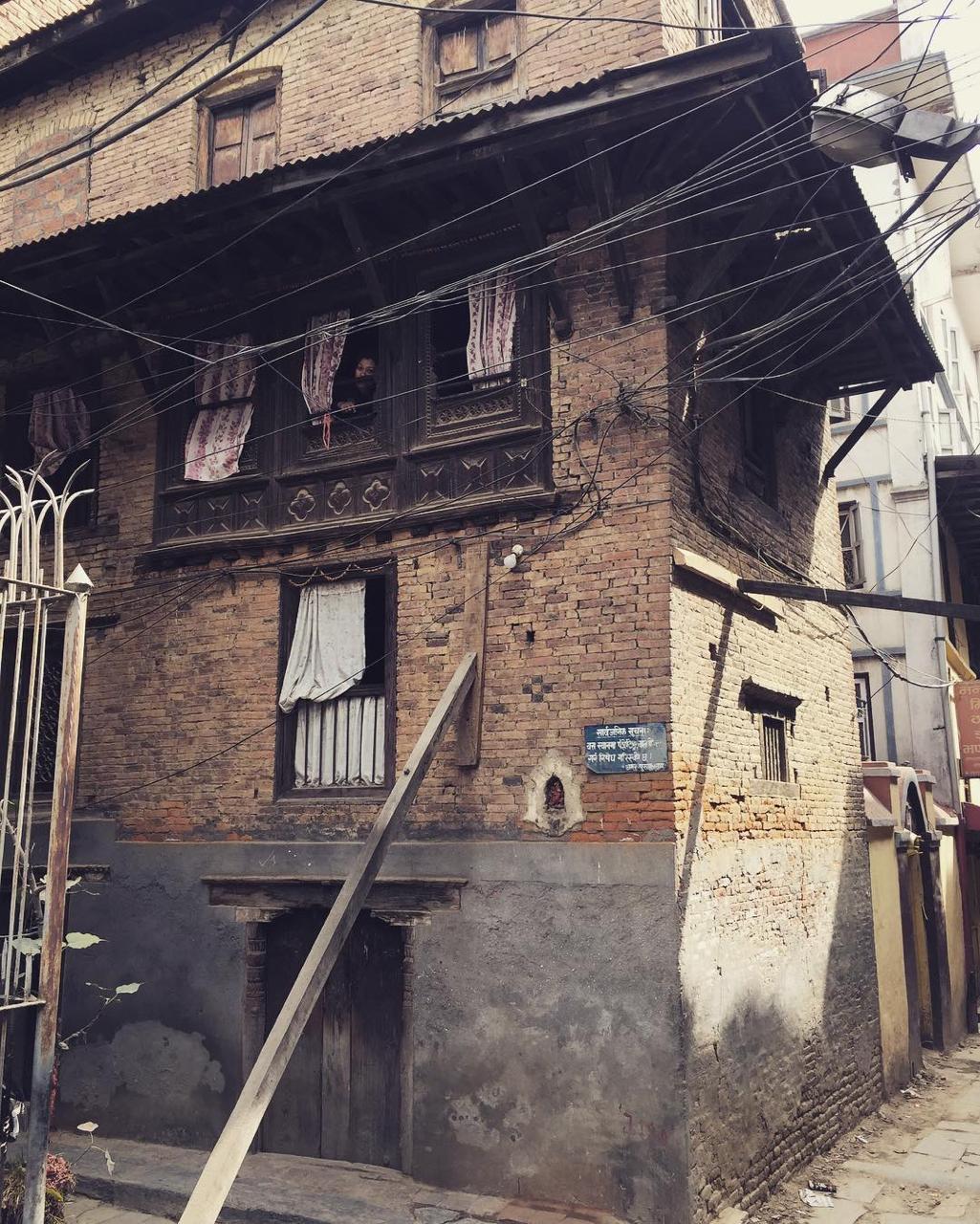}
    \end{subfigure}
    \begin{subfigure}{0.3\textwidth}
        \centering
        \includegraphics[width=\textwidth]{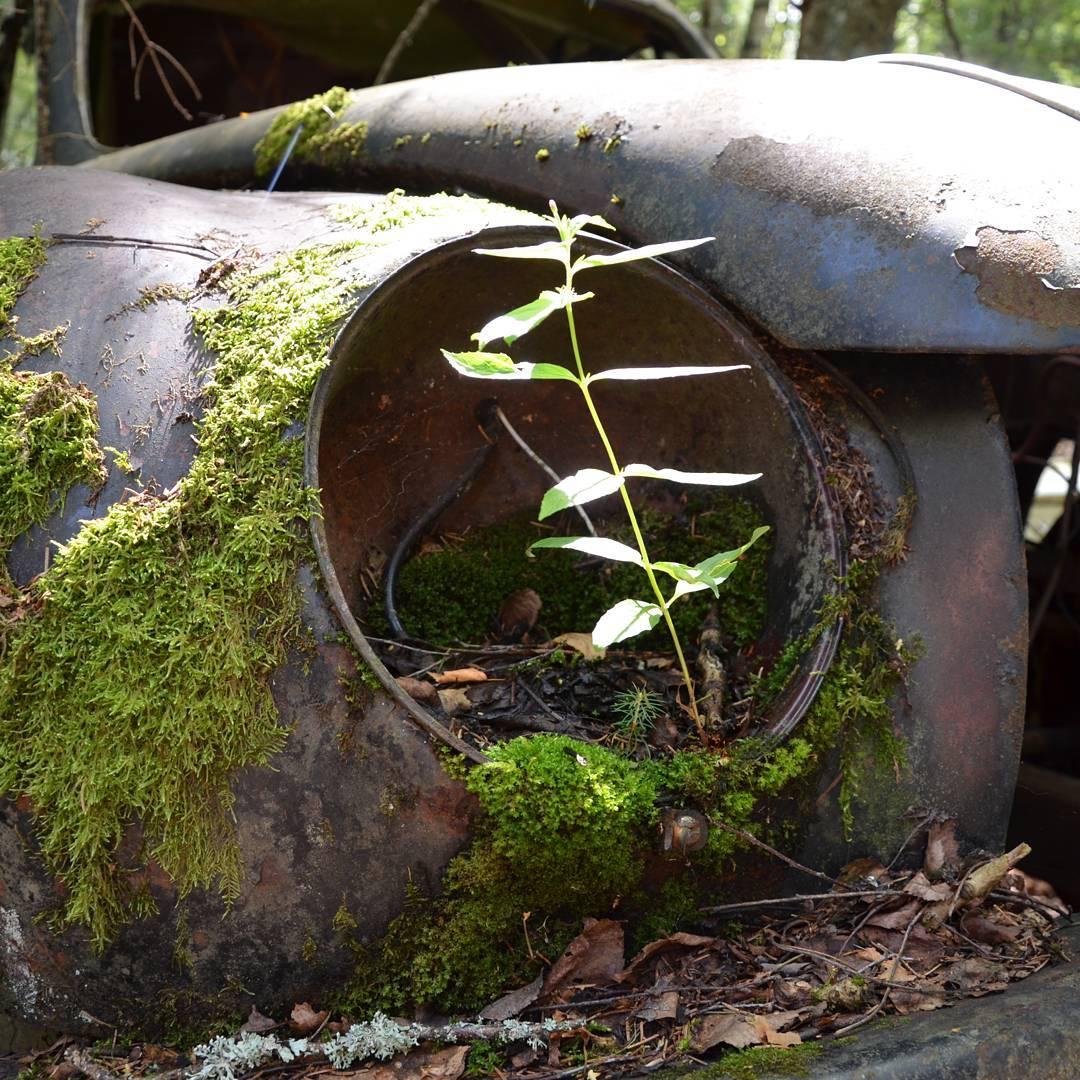}
    \end{subfigure}
    \caption{The error prediction by GPT-4o. In the first image, GPT-4o misclassified the event as a ``flood'' category while the input instance was tagged as an ``earthquake'' event. Similarly, for images 2 and 3, enough contextual signal is unavailable to deduce the correct event type.}
    \label{fig:performance_by_type_frequency}
    \vspace{-12pt}
\end{figure}
\begin{wrapfigure}{c}{0.5\linewidth}
\vspace{-15pt}
    \centering
    \includegraphics[width=\linewidth]{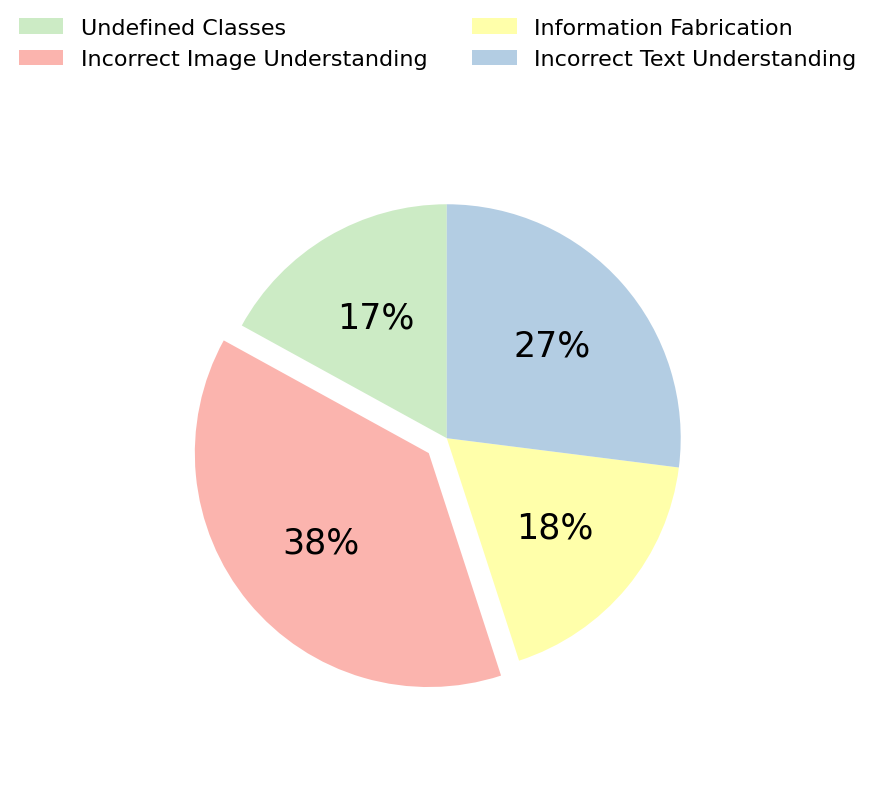}
    \caption{Error  Categorization of Generative Approaches.}
    \label{fig:error_cat}

\end{wrapfigure}

\vspace{-10pt}
\paragraph{\textbf{Error Analysis for Generative Approaches.}}
To systematically analyze the limitations of our models, we randomly sampled 250 erroneous predictions from both LLaVA and GPT-4o, categorizing them into four distinct error types. This categorization provides insights into the primary failure modes and informs future improvements to enhance robustness. As illustrated in Figure~\ref{fig:error_cat}, the majority of errors stem from incorrect handling of image-based cues (~38\%) and misinterpretation of textual information (~27\%). A key challenge arises in event classification scenarios where visual evidence alone is insufficient to disambiguate the event type. 
For instance, Figure~\ref{fig:performance_by_type_frequency} presents misclassified examples of ``earthquake'' events. All three images lack strong visual cues explicitly indicating an earthquake, making it difficult for even human annotators to confidently infer the correct event type. A particularly illustrative case is the left-most image, which depicts a river scene. The absence of visible structural damage or ground displacement makes it prone to misclassification as a ``flood'' event, highlighting a fundamental challenge in relying solely on visual data for event inference. Such cases emphasize the need for stronger multi-modal reasoning, where both textual and visual inputs are jointly contextualized to improve event classification accuracy.

We also observe some information mismatches in ED generations such as the prediction of classes not provided in the prompt (Undefined Classes) and the generation of sentences not provided in the desired output format (Information Fabrication). In the future, we can look into prompt design or prompt optimization \cite{ref24,ref26} to minimize such errors.  

\vspace{-5pt}
\section*{Limitations}
Our current study lays the foundation for effective disaster event identification using a multimodal analysis system. However, there are several avenues for future research and enhancements to our approach:

\begin{enumerate}
    \item \textbf{Advanced Prompt Design and Approaches:} We plan to explore advanced fusion techniques for integrating vision and language models, such as Contrastive Language-Image Pretraining (CLIP), LLaVa, and GPT-4o in few-shot setup with prompt optimization~\cite{ref24,ref26}. In addition, we will also explore instruction-tuning to boost the performance~\cite{ref27}.
    
    \item \textbf{Real-Time Monitoring and Response:} Integrating real-time monitoring capabilities into our system would enable immediate detection and response to unfolding disaster events. This involves developing mechanisms for continuous data scraping, processing, and analysis to provide timely alerts and actionable insights to response teams.
    
    \item \textbf{Scalability and Generalization:} As the volume and diversity of social media data continue to grow, ensuring the scalability and generalization of our system becomes crucial. Future work will focus on optimizing our models for scalability, adapting to evolving disaster event scenarios, and generalizing across different geographical regions and languages.
    
    \item \textbf{User-Generated Content Analysis:} Analyzing user-generated content, such as comments and replies to social media posts, can provide valuable contextual information and sentiment analysis related to disaster events. Incorporating this aspect into our analysis could enrich the understanding of public perceptions and responses during crises.
    
    \item \textbf{Collaborative Disaster Response Platforms:} Collaborative platforms that integrate our multimodal analysis system with other tools and resources used by disaster response teams could streamline coordination and decision-making processes. Future work will involve exploring integration possibilities and evaluating the impact on response effectiveness.
\end{enumerate}

By addressing these areas of future work, we aim to continually improve the capabilities and applicability of our multimodal analysis system for disaster event identification, ultimately contributing to more efficient and informed disaster response efforts.

\end{document}